\RequirePackage{fix-cm}
\documentclass[smallextended]{svjour3}       
\smartqed  
\usepackage{graphicx}
\usepackage{amsmath}

\usepackage{comment}
\usepackage[square,numbers]{natbib}
\usepackage{xcolor}
\usepackage{subfigure}
\usepackage{color}
\usepackage{multirow}
\usepackage{adjustbox}
\usepackage{colortbl}
\usepackage{siunitx}
\usepackage{amssymb}
\usepackage{algorithm}
\usepackage{algorithmic}
\usepackage[algo2e]{algorithm2e}
\begin{document}

\title{Efficient Scale Estimation Methods using Lightweight Deep Convolutional Neural Networks for Visual Tracking
}


\author{Seyed Mojtaba Marvasti-Zadeh \and
        Hossein Ghanei-Yakhdan  \and
        Shohreh Kasaei 
}

\institute{S. M. Marvasti-Zadeh \at
              Digital Image and Video Processing Lab (DIVPL), Department of Electrical Engineering, Yazd University, Yazd, Iran.\\
              Image Processing Lab (IPL), Department of Computer Engineering, Sharif University of Technology, Tehran, Iran.\\
              Vision and Learning Lab, Department of Electrical and Computer Engineering, University of Alberta, Edmonton, Canada.
              \\
              \email{mojtaba.marvasti@ualberta.ca}           
           \and
           H. Ghanei-Yakhdan (Corresponding Author) \at
              Digital Image and Video Processing Lab (DIVPL), Department of Electrical Engineering, Yazd University, Yazd, Iran.
              \\
              \email{hghaneiy@yazd.ac.ir}
           \and
           S. Kasaei \at
              Image Processing Lab (IPL), Department of Computer Engineering, Sharif University of Technology, Tehran, Iran.
              \\
              \email{kasaei@sharif.edu}
}

\date{Received: date / Accepted: date}

\maketitle

\begin{abstract}
In recent years, visual tracking methods that are based on discriminative correlation filters (DCF) have been very promising. However, most of these methods suffer from a lack of robust scale estimation skills. Although a wide range of recent DCF-based methods exploit the features that are extracted from deep convolutional neural networks (CNNs) in their translation model, the scale of the visual target is still estimated by hand-crafted features. Whereas the exploitation of CNNs imposes a high computational burden, this paper exploits pre-trained lightweight CNNs models to propose two efficient scale estimation methods, which not only improve the visual tracking performance but also provide acceptable tracking speeds. The proposed methods are formulated based on either holistic or region representation of convolutional feature maps to efficiently integrate into DCF formulations to learn a robust scale model in the frequency domain. Moreover, against the conventional scale estimation methods with iterative feature extraction of different target regions, the proposed methods exploit proposed one-pass feature extraction processes that significantly improve the computational efficiency. Comprehensive experimental results on the OTB-50, OTB-100, TC-128 and VOT-2018 visual tracking datasets demonstrate that the proposed visual tracking methods outperform the state-of-the-art methods, effectively.  
\end{abstract}

\section{Introduction}
\label{sec:1_Intro}
Visual tracking is a challenging research area in computer vision with a wide variety of practical applications, such as autonomous robots, driver assistance, video surveillance, and human-computer interaction \cite{SurveySmeulders,Yilmaz2006,SurveyDeepTracking}. These applications are faced with many challenging attributes (including heavy illumination variation, in-plane and out-of-plane rotations, partial or full occlusions, deformation, fast motion, camera motion, background clutter, viewpoint change, scale change, and so forth). Roughly speaking, visual tracking methods are mainly categorized into generative \cite{Kwon2014,Generative1,Generative2}, discriminative \cite{NeuralComputing1,NeuralComputing2,StruckTPAMI,RADSST}, and hybrid generative-discriminative \cite{Zhou2017,ASTCT,Liu2019} methods, which indicate to the appearance model of a visual target. In fact, generative methods search the best matching region with the learned target appearance and discriminative methods seek to estimate a decision boundary between the target and its background samples. While generative trackers suffer from discarding useful information of target surroundings, a large number of samples in discriminative trackers may lead to extensive computations. Hybrid generative-discriminative trackers benefit from both categories, but the exploration of generative and discriminative parts is a critical problem.

Due to the desired robustness and computational efficiency, DCF-based methods have extensively been used for visual tracking purposes. The aim of DCF-based methods is to learn a set of discriminative filters that an element-wise multiplication of them with a training set of feature maps in the frequency domain produces a desirable response. For localization step of visual tracking, many variants of DCF-based methods exploit either hand-crafted features (e.g., histogram of oriented gradients (HOG), histogram of local intensities (HOI), global color histogram (GCH), or Color Names (CN)), deep features from deep neural networks, or both. One of the most challenging attributes, in terms of tracking failures, is the scale changes of a visual target. Most of DCF-based visual trackers still exploit hand-crafted features for scale estimation step to preserve the computational efficiency although these features do not provide powerful scale models for a wide range of real-world applications. This paper aims to accurately estimate the scale of a visual target employing robust but extensive deep feature maps to improve the robustness of visual trackers against challenging scenarios and prevent drift problems. For preserving efficiency and improving the tracking performance of DCF-based visual tracking methods, this paper proposes the following contributions.

First, the exploitation of lightweight CNN models (including MobileNetV1 \cite{MobileNetV1} and MobileNetV2 \cite{MobileNetV2} models) in DCF-based visual tracking is proposed, which are highly efficient deep CNNs that are designed for mobile and embedded vision applications. Due to significantly fewer parameters and lower computational complexity, these pre-trained lightweight CNN models provide competitive visual tracking performance and acceptable tracking speed, simultaneously. 

Second, two efficient scale estimation methods for DCF framework-based methods are proposed, namely, holistic representation-based and region represe- ntation-based scale estimation methods. The proposed methods robustly estimate the target scale to prevent the contamination of target model with background information and cease the drift problem, effectively. 

Finally, the performance of proposed visual tracking methods is extensively compared with state-of-the-art visual trackers on four well-known visual tracking datasets.

The rest of the paper is organized as follows. The overview of related work including DCF-based visual trackers and their major scale estimation methods is described in Section \ref{sec:2_RelatedWork}. In Section \ref{sec:3_ProposedMethod}, the proposed holistic and region representation-based methods are presented. Comprehensive experimental results on different large visual tracking datasets are given in Section \ref{sec:4_ExpResults}, and finally, the paper is concluded in Section \ref{sec:5_Conclusion}.

\section{Related Work}
\label{sec:2_RelatedWork}
With the aid of dense sampling (i.e., circular data matrix) and efficient computations in the Fourier domain, the DCF-based methods have become one of the successful tracking-by-detection methods for visual tracking. In this section, a variety of DCF-based visual trackers is briefly explained. Then, three major scale estimation methods in these visual trackers are explored.

\subsection{DCF-based Visual Tracking Methods}
\label{sec:2_1}
The popularity of DCF-based visual trackers goes back to the minimum output sum of squared error (MOSSE) correlation filter \cite{MOSSE}, which was a fast and accurate tracker against complex traditional visual tracking methods. In this tracker, the target location is estimated by the maximum value of a confidence map after correlating the learned filter from a given target over a search region in the next frame. To address the computational burden of DCF-based visual trackers (to increase limited training samples by cyclically shifted samples), the exploitation of circulant matrices was proposed \cite{KCF-ECCV} that formulates the classical DCF-based methods as a ridge regression problem. Based on this framework, called dense sampling, the first single-channel kernelized DCF-based visual tracker with a fast and closed-form solution was proposed. It uses the fast Fourier transform to achieve extremely fast learning and detecting steps. With the aid of discrete Fourier transform, the multi-channel kernelized correlation filter (KCF) \cite{KCF} diagonalizes the circulant data matrix to not only reduce both storage and computational cost but also to exploit multi-channel features (e.g., HOG). Based on kernel ridge regression, the KCF does not suffer from the curse of kernelization and pleasantly runs at hundreds of frames-per-second (FPS) that is desired for real-time applications. 

The correlation filters with limited boundaries \cite{CFLimitedBoundaries} aim to drastically limit the unwanted boundary effects, which are created by cyclically shifted training samples. This visual tracker iteratively solves a slightly augmented objective function to not only reduce the dramatic boundary effect on the performance but also to preserve the computational advantages of the dense sampling strategy. Moreover, the spatially regularized discriminative correlation filters (SRDCF)-based method \cite{SRDCF} introduces a spatial regularization component to the standard DCF formulation to improve the robustness of the appearance model. Moreover, the context-aware correlation filters \cite{BACF} with the purpose of resolving conventional DCF formulations exploit additional context information (i.e., immediate spatial and temporal backgrounds of target) to better discriminate between target and background.

To further improve the target appearance model, deep features (mostly achieved by CNNs) have demonstrated remarkable performance in DCF-based visual tracking methods. To effectively pose the DCF formulation in the continuous spatial domain, the continuous convolution operator tracker (C-COT) \cite{CCOT} utilizes an implicit interpolation method to learn continuous convolution filters. It then fuses multi-resolution feature maps (e.g., multiple layers of deep feature maps, HOG, and CN) to construct a continuous confidence map. From the three perspectives of model size, training set size, and model update, the efficient convolution operators (ECO) tracker \cite{ECO} reformulates the C-COT by using a factorized convolution operator, compact generative model, and efficient model update strategy, respectively. These key factors reduce the over-fitting risk of model leading to enhance performance and computational efficiency, simultaneously. 

The hierarchical correlation features-based tracker (HCFT) \cite{HCFT} exploits hierarchical feature maps of CNNs to employ both sematic and fine-grained details for visual tracking. Also, a set of linear DCFs is adaptively learned on different CNN layers to determine the target location in a coarse-to-fine fashion. Also, the advanced version of HCFT (HCFTs or HCFT*) \cite{HCFTs} not only benefits from the HCFT advantages but also learns another DCF with long-term memory of target appearance to handle critical situations; such as heavy occlusion or out-of-view. With the aid of three types of correlation filters (including translation filter, scale filter, and long-term filter), the long-term DCF-based tracker, called LCT \cite{LCTdeep}, learns a long-term complementary DCF after estimating the position and scale of a visual target. The long-term complementary filter determines the tracking failures. Then, an incrementally learned detector recovers the visual target in a sliding window fashion. While the LCT tracker benefits hand-crafted features, the other version of this tracker, called LCTdeep \cite{LCTdeep}, exploits both hand-crafted and deep features, simultaneously.

\subsection{Scale Estimation Methods in DCF-based Trackers}
\label{sec:2_2}
Most of state-of-the-art DCF-based visual trackers employ one of three scale estimation methods, including the multi-resolution translation filter, a joint scale-space filter, or discriminative scale space tracking \cite{DSST}, which are briefly described as follows.

As a conventional method in object detection, the multi-resolution translation filter firstly learns a 2D standard DCF for a given visual target. Then, the learned DCF is applied on multiple extracted patches (i.e., target candidates) with different resolutions to select the best candidate with the highest correlation score. The joint scale-space filter is the other straightforward method to estimate the target scale. With the aid of a 3D Gaussian function (as the desirable correlation output) and after constructing a 3D region (which includes two spatial dimensions and one scale dimension in each level) by using a feature pyramid around a given target, the joint scale-space filter estimates both translation and scale by finding the best correlation score.

Despite the aforementioned methods for target scale estimation, the discriminative scale space tracking (DSST) \cite{DSST} applies two independent translation and scale correlation filters for robust visual tracking. First, the DSST extracts the HOG feature maps for several patches of a given target at different resolutions. Second, it reshapes the feature maps to the 1D feature descriptors to be able to exploit a 1D Gaussian function as the desired output. Third, 1D feature descriptors are concatenated to construct a 2D scale DCF. Finally, the scale DCF is used for accurate and robust scale estimation with the same formulations as the translation DCF.

Although deep neural networks have led to many significant advances in visual tracking methods (e.g., robust target representation), the state-of-the-art DCF-based visual trackers still exploit hand-crafted features (e.g., HOG or its variants like PCA-HOG) for scale estimation \cite{CCOT,ECO,STRCF}. As it is evident from the explained scale estimation methods, several target feature extractions are needed for different target resolutions. Therefore, the multiple exploitation of deep neural networks can lead to a heavy computational burden. To address the limitation of using deep neural networks for robust scale estimation, two-scale estimation methods using lightweight CNNs are proposed. The proposed visual tracking methods construct robust translation and scale models using lightweight MobileNet models to provide an acceptable computational cost for visual target tracking methods.

\section{Proposed Visual Tracking Methods}
\label{sec:3_ProposedMethod}
In this section, two visual tracking methods based on efficient convolution operators \cite{ECO} are proposed. The proposed methods exploit state-of-the-art MobileNet models in the DCF-based visual tracking to robustly estimate the target scale. The proposed methods utilize a translation model same as \cite{ECO} with the objective function in the Fourier domain as \cite{SRDCF}
\begin{equation}\label{Eq.1}
{\mathop{\mathrm{min}}_{H,C}\ \  {\left\|{\bar{X}}^TC\bar{H}-\bar{Y}\right\|}^2_{{\ell }^2}+\sum^m_{i=1}{{\left\|\bar{S}\circledast {\bar{H}}^i\right\|}^2_{{\ell }^2}+}\lambda {\left\|C\right\|}^2_{\mathbb{F}}\ }
\end{equation}
in which $X$, $C$, $H$, $Y$, $S$, $m$, and $\lambda$ denote the interpolated feature maps from MobileNet models, linear dimensionality reduction matrix, multi-channel convolution filters, desirable response map, spatial penalty, number of a smaller set of basis filters, and regularization parameter, respectively. Moreover, the L2 and Frobenius norms are denoted by ${\ell}^2$ and $\mathbb{F}$. Also, the $\circledast$ and bar symbols represent the multi-channel convolution operator and Fourier series coefficients of variables, respectively. To optimize the objective function, the Gauss-Newton and Conjugate Gradient methods are used \cite{ECO}. 

In the following, the proposed efficient scale estimation methods are explained. The proposed methods using lightweight CNN models learn a 2D scale correlation filter according to the holistic or region-based representation of convolutional feature maps.

\subsection{Proposed Holistic Representation-based Scale Estimation Method}
\label{sec:3_1}
Generally, most of the deep visual tracking methods (which exploit deep feature maps from pre-trained CNN models) have been trained on large-scale image datasets (e.g., ImageNet dataset \cite{ImageNet}). The learned filters of these CNN models provide the holistic representation of a visual scene. The proposed holistic representation-based scale estimation method (HRSEM) exploits this characteristic of deep feature maps to robustly estimate the target scale. The HRSEM exploits the one-pass feature extraction process of each full video frame through the network to learn a 2D scale correlation filter. Given the 3D deep feature maps (denoted by $F$), the interpolated cropped feature maps ${W_{PF}}$ are defined as
\begin{equation}\label{Eq.2}
W_{PF}=PF*{V_F}
\end{equation}
in which ${V_F}$ is the interpolation model used in \cite{ECO}, given by
\begin{equation}\label{Eq.3}
V_F\left(t\right)=\sum^{R-1}_{n=1}{{\mathbb{H}}_{sc}\left[n\right]K\left(t-\frac{T}{R}n\right)}
\end{equation}
where $H_{sc}$, $K$, and $R$ are the multi-channel scale correlation filters (i.e., scale model), $T$-periodic interpolation kernel, and independent resolution of each feature map, respectively. The interpolation model transfers deep feature maps to the continuous spatial domain $t$ to provide an accurate scale estimation. The tensor $P$ includes $G$ binary matrices (which $G$ is the number of feature maps) with the size $M\times N$ (i.e., the same size to the spatial dimensions of video frames) to crop target regions at different scales $D$. Given an $L\times U$ target dimensions, the pre-defined scale set of a visual target is specified by
\begin{equation}\label{Eq.4}
\mathcal{Q}=\left\{a^bL\times a^bU\ \mathrel{\left|\vphantom{a^bL\times a^bU\  \ b\in \left(\left\lfloor {-(D-1)}/{2}\right\rfloor ,\dots ,\left\lfloor {(D-1)}/{2}\right\rfloor \right)}\right.\kern-\nulldelimiterspace}\ b\in \left(\left\lfloor {-(D-1)}/{2}\right\rfloor ,\dots ,\left\lfloor {(D-1)}/{2}\right\rfloor \right)\right\}
\end{equation}
in which $a$, $b$, and $D$ are the scale factor, scale level, and size of scale filters, respectively. To enjoy the computation efficiency of the DCF formulations, the vectorization of interpolated cropped feature maps $W_{PF}$ (with $M\times N\times G$ dimension) for each pre-defined scale is defined as
\begin{equation}\label{Eq.5}
\overrightarrow{W_{PF}}={(J^T_{1,1},\dots ,J^T_{N,1},\dots ,J^T_{1,G},\dots ,J^T_{N,G})}^T
\end{equation}
where $\overrightarrow{W_{PF}}$, $J_{\mathfrak{i},\mathfrak{j}}$, and \textit{${}^{T}$} represent the vectorized deep features, $i$-th colomn of $W_{PF}$ at $j$-th deep feature channel, and transpose operator, respectively. Then, the multi-scale deep feature map $\mathbb{W}$ is achieved by 
\begin{equation}\label{Eq.6}
\mathbb{W}=\left[\overrightarrow{W^1_{PF}},\overrightarrow{W^2_{PF}},\dots ,\overrightarrow{W^D_{PF}}\ \right].
\end{equation}
To learn scale correlation filters, the proposed scale estimation methods utilize the following objective function (i.e., ridge regression) in the spatial domain 
\begin{equation}\label{Eq.7}
{\mathop{\mathrm{min}}_{H_{sc}}\ \  {\left\|{\mathbb{Y}}_{sc}-\mathbb{W}{\mathbb{H}}_{sc}\right\|}^2_{{\ell }^2}+\lambda {\left\|{\mathbb{H}}_{sc}\right\|}^2_{{\ell }^2}\ }
\end{equation}
where subscript $SC$ indicates the variables corresponding to the proposed scale correlation filters, and $\mathbb{Y}_{sc}$ is the desirable correlation response, respectively. The closed-form learning and detection solutions \cite{KCF} (in the Fourier domain) are respectively applied by 
\begin{equation}\label{Eq.8}
{\bar{\mathbb{H}}}_{sc}=\frac{{\bar{\mathbb{W}}}^*\odot {\bar{\mathbb{Y}}}_{sc}}{{\bar{\mathbb{W}}}^*\odot \bar{\mathbb{W}}+\lambda}
\end{equation}
\begin{equation}\label{Eq.9}
\mathbb{E}=\sum^G_{g=1}{{\bar{\mathbb{H}}}^{t-1}_{sc,g}\odot {\bar{\mathbb{Z}}}^t_g}
\end{equation}
where $\mathbb{E}$, $\mathbb{H}^{t-1}_{sc,g}$, ${\mathbb{Z}}^t$, $^*$, and $\odot$ represent the scale confidence map, $g$-th scale correlation response achieved by previous frames, vectorized features of image patch (i.e., search region) in the current frame (which is centered around the previous location of the target), complex-conjugate operator, and element-wise product, respectively. By using the standard Newton's method, the target scale is iteratively estimated by finding the maximum value of $\mathbb{E}$ in \ref{Eq.9}. Finally, the scale model is updated by the convenient update rule, given by
\begin{equation}\label{Eq.10}
{\mathbb{H}}^t_{sc}=\left(1-\eta \right){\mathbb{H}}^{t-1}_{sc}+\eta {\mathbb{H}}^t_{sc}
\end{equation}
in which $\eta $ denotes a learning rate parameter.

\subsection{Proposed Region Representation-based Scale Estimation Method}
\label{sec:3_2}
To avoid the iterative feature extraction process, the region representation-based scale estimation method (RRSEM) is proposed. Similar to the translation model that exploits deep features of the target region, the proposed RRSEM utilizes a batch of target regions at different scales to estimate target scale using a CNN model. 

For a visual target with \textit{L}$\times $\textit{U} dimensions, the proposed method employs the pre-defined scale set $\mathcal{Q}$ to construct a set of target regions at different scales (denoted by $I$). The interpolated target regions ${\mathrm{\Psi }}_{\mathrm{i}}$ with the size of $M\times N\times \mathbb{C}$ (where $\mathbb{C}$ is the number of frame channels) are achieved by
\begin{equation}\label{Eq.11}
{\mathrm{\Psi }}_i=I_i*V_F\ \ \ ,\ \ \ \ \ i\ \in \left\{1,\dots ,\ D\right\}\ .
\end{equation}
Then, the proposed RRSEM concatenates the regions ${\mathrm{\Psi }}_i$ by
\begin{equation}\label{Eq.12}
\mathbb{I}=\left[{\mathrm{\Psi }}_1,\dots ,{\mathrm{\Psi }}_D\right]
\end{equation}
to construct a 4D batch of target regions (denoted by $\mathbb{I}$).
Thanks to the batch of target regions, the proposed RRSEM can extract 4D deep feature maps $F$ (with $M^{'}\times N^{'}\times G\times D$ dimensions) that include different scales of the target. Afterward, the vectorized deep features $\overrightarrow{F_b}$ (at the $b$ scale level) are achieved by
\begin{equation}\label{Eq.13}
\overrightarrow{F_b}={(F^T_{1,1,b}\ ,\dots ,F^T_{N^{'},1,b}\ ,\dots ,F^{T}_{1,G,b}\ ,\dots ,F^{T}_{N^{'},G,b})}^{T}.
\end{equation}
Then, the multi-scale deep feature map $\mathbb{W}$ for the proposed method can be achieved by
\begin{equation}\label{Eq.14}
\mathbb{W}=\left[\overrightarrow{F_1},\overrightarrow{F_2},\dots ,\overrightarrow{F_D}\ \right].
\end{equation}
Finally, the proposed region representation-based method uses \ref{Eq.7} to \ref{Eq.10} to learn and estimate the scale of a given visual target. Algorithm \ref{alg:PropMethods} represents the learning and detection processes of proposed methods for scale estimation.

\begin{algorithm}
\caption{Proposed Scale Estimation Methods for DCF Framework}
\label{alg:PropMethods}
\textbf{Input:} Sequence frames ($T$: sequence length), Initial bounding box of target (i.e., target region) \leavevmode \\
\textbf{Output:} Estimated target scales \leavevmode \\
 \For{$t=1:T$}{
    \If{$t>1$}{ \leavevmode \\
       Extract search region features by MobileNetV2 \leavevmode \\
       Vectorize features by Eq. (5) \leavevmode \\
       Apply FFT on vectorized features \leavevmode \\
       Compute scale confidence map (CM) by Eq. (9) \leavevmode \\
       Estimate target scale by finding maximum value of scale CM
    }
    \If{{method: HRSEM}}{ \leavevmode \\
       Extract features of frame $t$ by MobileNetV2 \leavevmode  \\
       Vectorize features by Eq. (5) \leavevmode \\
       Construct multi-scale feature maps by (6) \leavevmode \\
       Apply FFT on multi-scale feature maps \leavevmode \\
       Learn 2D scale CFs by Eq. (8) 
    }
    \If{{method: RRSEM}}{ \leavevmode \\
       Extract target regions at different scales \leavevmode \\
       Compute interpolated target regions by Eq. (11) \leavevmode \\
       Concatenate interpolated regions by Eq. (12) \leavevmode \\
       Extract features of concatenated regions by MobileNetV2 \leavevmode \\
       Vectorize features by Eq. (13) \leavevmode \\
       Construct multi-scale feature maps by (14) \leavevmode \\
       Apply FFT on multi-scale feature maps \leavevmode \\
       Learn 2D scale CFs by Eq. (8) 
    }
    \If{$t>1$}{ \leavevmode \\
       Update scale model by Eq. (10) 
    }
 }
\end{algorithm}
\section{Experimental Results}
\label{sec:4_ExpResults}
This section provides the implementation details, baseline comparison (which indicates the impact of exploiting lightweight CNN models in the proposed methods), and the performance comparisons with the state-of-the-art visual tracking methods, respectively. 

\subsection{Implementation Details}
\label{sec:4_1}
The baseline comparisons of proposed visual tracking methods are evaluated using the MobileNetV1 and MobileNetV2 models on the OTB-50 dataset \cite{OTB2013}. Then, the proposed visual tracking methods using the MobileNetv2 model are selected and compared with the state-of-the-art visual tracking methods on four well-known visual tracking datasets of OTB-50, OTB-100 \cite{OTB2015}, TC-128 \cite{TC128}, and VOT-2018 \cite{VOT-2018}, comprehensively (see Table \ref{Table_Datasets}). These datasets contain 51, 100, and 129 video sequences, respectively. They contain common challenging attributes including illumination variation (IV), scale variation (SV), fast motion (FM), out-of-view (OV), background clutter (BC), out-of-plane rotation (OPR), occlusion (OCC), deformation (DEF), motion blur (MB), and in-plane rotation (IPR), and low resolution (LR). Besides, the well-known precision and success plots \cite{OTB2013,OTB2015} are used to evaluate the performance of visual tracking methods. While the success plot shows how many percentages of frames have an overlap score more than a specific threshold ($50\%$ in this work) between their estimated and ground-truth bounding boxes, the percentage of frames with location estimation error smaller than a pre-defined threshold ($20$ pixels in this work) is shown by the precision plot. Moreover, the proposed methods is implemented on an Intel I$7-6800$K $3.40$ GHz CPU, $64$ GB RAM, with an NVIDIA GeForce GTX $1080$ GPU. 

\begin{table}[!bp]
\caption{Exploited Visual Tracking Datasets. [NoV: Number of Videos, NoF: Number of Frames].} 
\centering 
\resizebox{\textwidth}{!}{
\begin{tabular}{c c c c c c c c c c c c c c} \hline \hline 
Dataset & NoV & NoF & \multicolumn{11}{c}{NoV Per Attribute} \\ \hline 
 &  &  & IV & OPR & SV & OCC & DEF & MB & FM & IPR & OV & BC & LR \\ \hline 
OTB-50 & 51 & 29491 & 25 & 39 & 28 & 29 & 19 & 12 & 17 & 31 & 6 & 21 & 4 \\ \hline
OTB-100 & 100 & 59040 & 38 & 63 & 65 & 49 & 44 & 31 & 40 & 53 & 14 & 31 & 10 \\ \hline 
TC-128 & 129 & 55346 & 37 & 73 & 66 & 64 & 38 & 35 & 53 & 59 & 16 & 46 & 21 \\ \hline 
VOT-2018 & 70 & 25504 & \multicolumn{11}{c}{Frame-based attributes} \\ 
\hline\hline
\end{tabular}}
\label{Table_Datasets}
\end{table}
Similar to various trackers \cite{CCOT,ECO,HCFTs} that exploit the combination of deep feature maps from different layers of a CNN, the translation model of the proposed methods exploits the combination of “conv-2” and “conv-3” layers for MobileNetV1, and the combination of “conv-2” and “conv-4” layers for MobileNetV2. Moreover, the proposed scale estimation methods exploit the “conv-3” layer of MobileNetV2 model as the deep features to provide an appropriate balance of spatial and semantic information. These feature layers are selected experimentally. The other parameters of the translation model in the proposed methods are the same as \cite{ECO}. For fair and clear performance evaluations, all proposed methods merely exploit deep features. For the scale model, the proposed methods use $\eta = 0.025$, $a=1.02$, $D=17$, and the standard deviation of desirable correlation response and the number of Newton iterations are set to $1.0625$, and $5$, respectively. 

\subsection{Baseline Comparison}
\label{sec:4_2}
\begin{figure}
\centering
\includegraphics[width=8cm, height=8cm]{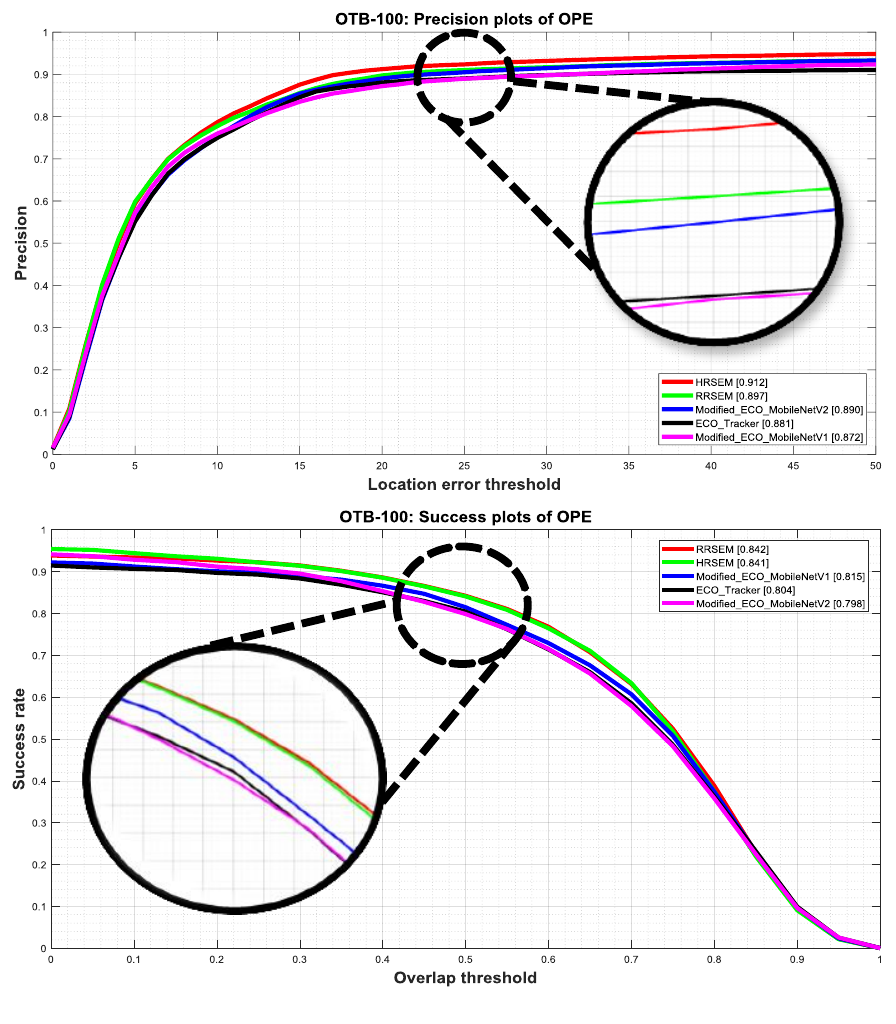}
\caption{Precision and success evaluations of baseline visual tracking methods on OTB-50 dataset.}\label{Fig1_BaseComp}
\vspace{-4mm}
\end{figure}

To demonstrate the efficiency of visual tracking with lightweight CNNs, the performance of two versions of MobileNet models (i.e., MobileNetV1 \cite{MobileNetV1} and MobileNetV2 \cite{MobileNetV2}) are evaluated on the OTB-50 dataset. On one hand, two modified versions of \cite{ECO} that exploit the fusion of either MobileNetV1 deep feature maps (namely, modified-ECO-MobileNetV1) or MobileNetV2 deep feature maps (namely, modified-ECO-MobileNetV2) are proposed. These proposed methods only use a different CNN model in their translation model and utilize hand-crafted features (i.e., PCA-HOG same as \cite{ECO}) for their scale model. As shown in Fig. \ref{Fig1_BaseComp}, the exploitation of MobileNet models have competitive visual tracking performance with \cite{ECO} that employs the VGG-M model \cite{VGGM}. However, the MobileNets can accelerate the process of deep feature extraction due to tremendously fewer parameters of their CNN models. For example, the tracking speeds of the proposed tracking methods using MobileNetV2 and MobileNetV1 models and also \cite{ECO} on the aforementioned machine are about $20$ FPS, $16$ FPS, and $10$ FPS, respectively. According to the higher efficiency of the MobileNetV2, this model is adopted as the best baseline method to integrate with the proposed scale estimation methods. 

On the other hand, two visual tracking methods are proposed that exploit the MobileNetV2 model in both translation and scale models (i.e., HRSEM and RRSEM). As shown in Fig. \ref{Fig1_BaseComp}, the proposed holistic and region representation-based visual tracking methods have improved the average precision rate up to $3.1\%$, and $1.6\%$, and the average success rate up to $3.7\%$, and $3.8\%$ compared to the ECO tracker \cite{ECO}, respectively. It is evident that the proposed scale estimation methods have clearly improved the performance of the baseline Modified-ECO-MobileNetV2 tracker while preserving the computational efficiency ($\sim 15$ and $12$ FPS for HRSEM and RRSEM, respectively). Although the experimental results on the OTB-50 dataset demonstrate that the MobileNetV2 model improves both target localization (i.e., precision metric) and visual tracking speed, it has slightly lower performance according to success metric. In the following, the performance of the proposed visual trackers (i.e., HRSEM and RRSEM) is compared with state-of-the-art visual tracking methods.

\subsection{Performance Comparison}
\label{sec:4_3}
To quantitatively evaluate the performance of visual trackers, the proposed visual tracking methods (i.e., HRSEM and RRSEM) are compared with $19$ visual trackers on the OTB-50 \cite{OTB2013} and OTB-100 \cite{OTB2015} datasets, $11$ visual trackers on the TC-128 dataset \cite{TC128}, and $20$ trackers on the VOT-2018 dataset \cite{VOT-2018} (for which the benchmark results are publicly available). Although the OTB-50, OTB-100, and TC-128 use the one-pass evaluation to compare the visual tracking methods, the VOT-2018 employs the visual tracking exchange protocol (TraX) \cite{TraX} that not only detects the tracking failures but also re-initializes visual trackers five frames after each failure. While the OTB-50, OTB-100, and TC-128 rank the visual trackers according to their area under curve (AUC) score, the VOT-2018 ranks these methods based on the mean of accuracy-robustness (AR) performance, which the smaller robustness score shows the more failures. 

The state-of-the-art visual tracking methods include wide range of deep feature-based trackers (including the LCTdeep \cite{LCTdeep}, UCT \cite{UCT}, ECO \cite{ECO}, HCFT \cite{HCFT}, HCFTs \cite{HCFTs}, DeepSRDCF \cite{DeepSRDCF}, SiamFC-3s \cite{SiamFC}, DCFNet \cite{DCFNet}, DCFN-et-2.0 \cite{DCFNet}, CCOT \cite{CCOT}, DSiam \cite{DSiam}, CFNet \cite{CFNet}, DeepCSRDCF \cite{DeepCSRDCF}, MCPF \cite{MCPF}, TRACA \cite{TRACA}, DeepSTRCF \cite{STRCF}, SiamRPN \cite{SiamRPN}, SA-Siam \cite{SA-Siam}, LSART \cite{LSART}, DRT \cite{DRT}, DAT \cite{DAT}, C-RPN \cite{CRPN}, GCT \cite{GCT}, SiamDW-SiamRPN \cite{SiamDW}, and SiamDW-SiamFC \cite{SiamDW}) that exploit deep neural networks as either a feature extraction network or an end-to-end network. Also, diverse hand-crafted feature-based trackers (including the LCT \cite{LCTdeep}, SRDCF \cite{SRDCF}, MUSTer \cite{MUSTer}, DSST \cite{DSST}, Staple \cite{Staple}, MEEM \cite{MEEM}, BACF \cite{BACF}, SAMF \cite{SAMF}, KCF \cite{KCF}, and Struck \cite{StruckTPAMI}) are used that exploit different features (including HOG, HOI, GCH, CN, scale-invariant feature transform (SIFT), Haar, and raw pixels) and strategies for visual tracking.

\begin{figure}
\centering
\includegraphics[width=11.9cm, height=13cm]{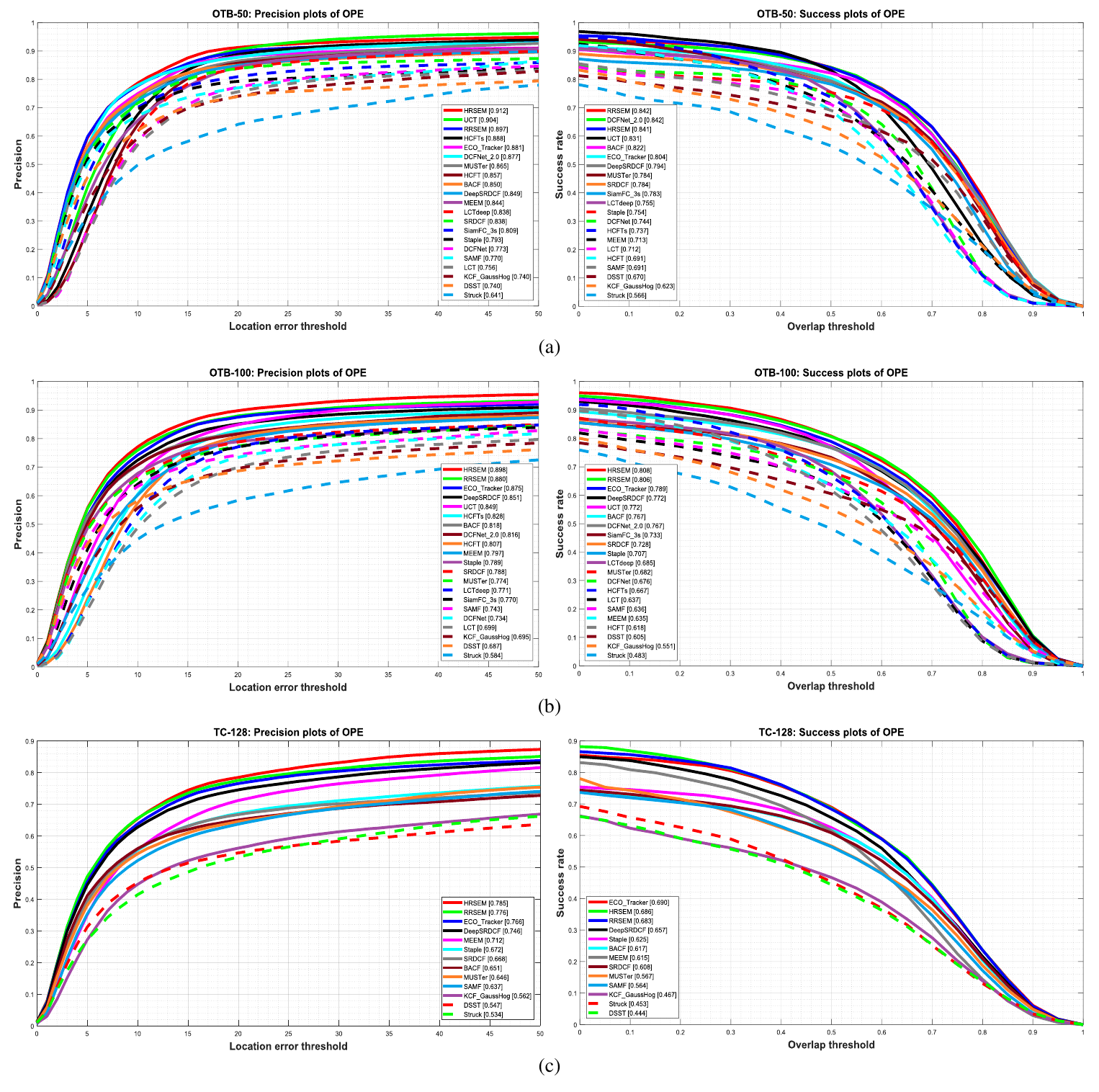}
\caption{Overall precision and success evaluations of proposed methods using MobileNetV2 compared with state-of-the-art visual trackers on large visual tracking datasets: (a) OTB-50, (b) OTB-100, (c) TC-128.}\label{Fig2_OverallComp}
\vspace{-4mm}
\end{figure}

The precision and success plots of visual tracking methods on the OTB-50, OTB-100, and TC-128 datasets are shown in Fig. \ref{Fig2_OverallComp}. These results indicate that the HRSEM has improved the average precision rate up to $0.8\%$, $2.4\%$, $3.5\%$, $3.1\%$, $6.2\%$, and $6.3\%$, and the average success rate up to $1\%$, $10.4\%$, $-0.1\%$, $3.7\%$, $1.9\%$, and $4.7\%$ compared to the UCT, HCFTs, DCFNet-2.0, ECO, BACF, and DeepSRDCF on the OTB-50 dataset, respectively. The experimental results of the HRSEM with aforementioned visual trackers on the OTB-100 dataset, show that the average precision rate up to $4.9\%$, $7\%$, $8.2\%$, $2.3\%$, $8\%$, and $4.7\%$, and the average success rate up to $3.6\%$, $14.1\%$, $4.1\%$, $1.9\%$, $4.1\%$, and $3.6\%$ has improved, respectively. 

Finally, the obtained results show that the HRSEM has increased the average precision rate up to $13.4\%$, $3.9\%$, and $1.9\%$, the average success rate up to $6.9\%$, $2.9\%$, and $-0.4\%$, on the TC-128 dataset compared to the BACF, DeepSRDCF, and ECO trackers, respectively. As shown in Fig. \ref{Fig2_OverallComp}, the proposed visual tracking methods outperform the state-of-the-art visual tracking methods in terms of precision and success rates on visual tracking datasets. The success of proposed methods is due to the efficient exploitation of the lightweight CNN model and the integration of deep feature maps in the DCF formulations. The building blocks of the MobileNetV2 model are based on bottleneck depth-separate convolution with residuals that significantly reduce the computational cost for extracting deep feature maps. 

\begin{figure}
\centering
\includegraphics[width=11.9cm, height=3.5cm]{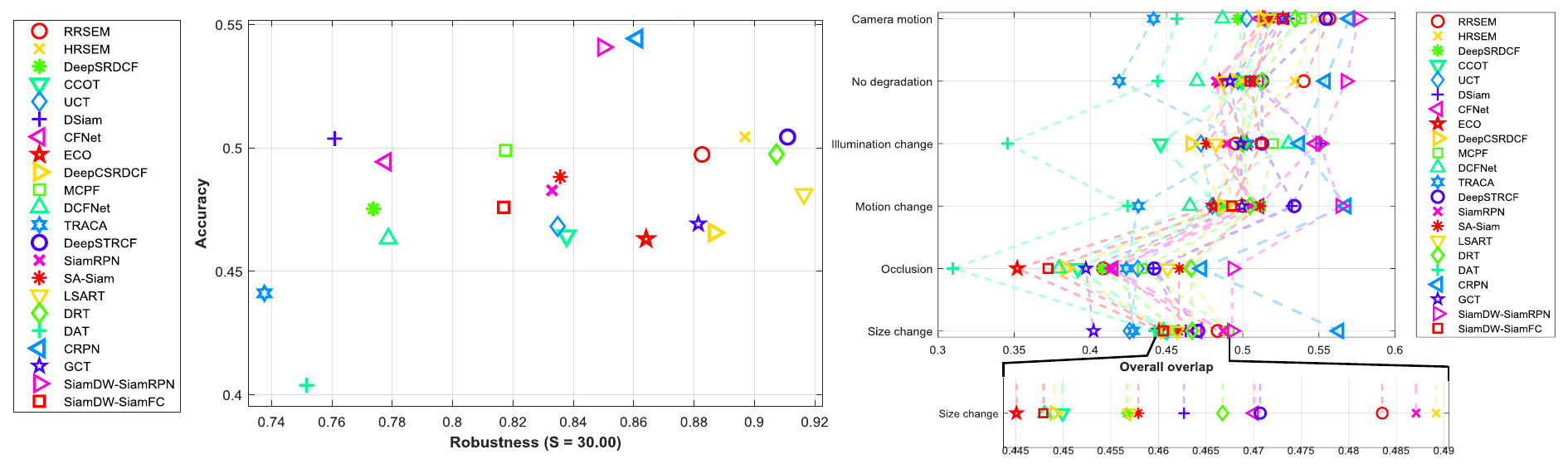}
\caption{Overall and attribute-based comparisons of proposed methods using MobileNetV2 compared with state-of-the-art visual trackers on VOT-2018 dataset.}\label{Fig3_VOTComp}
\vspace{-4mm}
\end{figure}

Utilizing recent research on designing smaller CNNs with desirable representational power, the proposed methods construct robust translation and scale models of a visual target using deep feature maps for DCF-based tracking. In comparison with models that exploit hand-crafted features, these models help the proposed methods to appropriately handle the challenging visual tracking scenarios. Moreover, the proposed formulations suitably facilitate the integration of deep feature maps into the DCF formulations, which obtain high computational efficiency in the Fourier domain. Although the proposed visual tracking methods have slightly lower success rate on the TC-128 dataset compared to \cite{ECO} (which is because of the lower success rate of the MobileNetV2 model compared to the VGG-M model (see Fig. \ref{Fig1_BaseComp}) for visual tracking purposes), the proposed methods have beneficially increased the general visual tracking performance without a high computational burden. Fig. \ref{Fig3_VOTComp} shows the overall and attribute-based comparisons of the proposed methods with the state-of-the-art visual tracking on the VOT-2018 dataset. The achieved results illustrate that the proposed methods improve the accuracy and robustness of visual tracking compared to the baseline tracker \cite{ECO}. 

\begin{figure}
\centering
\includegraphics[width=11.9cm, height=9cm]{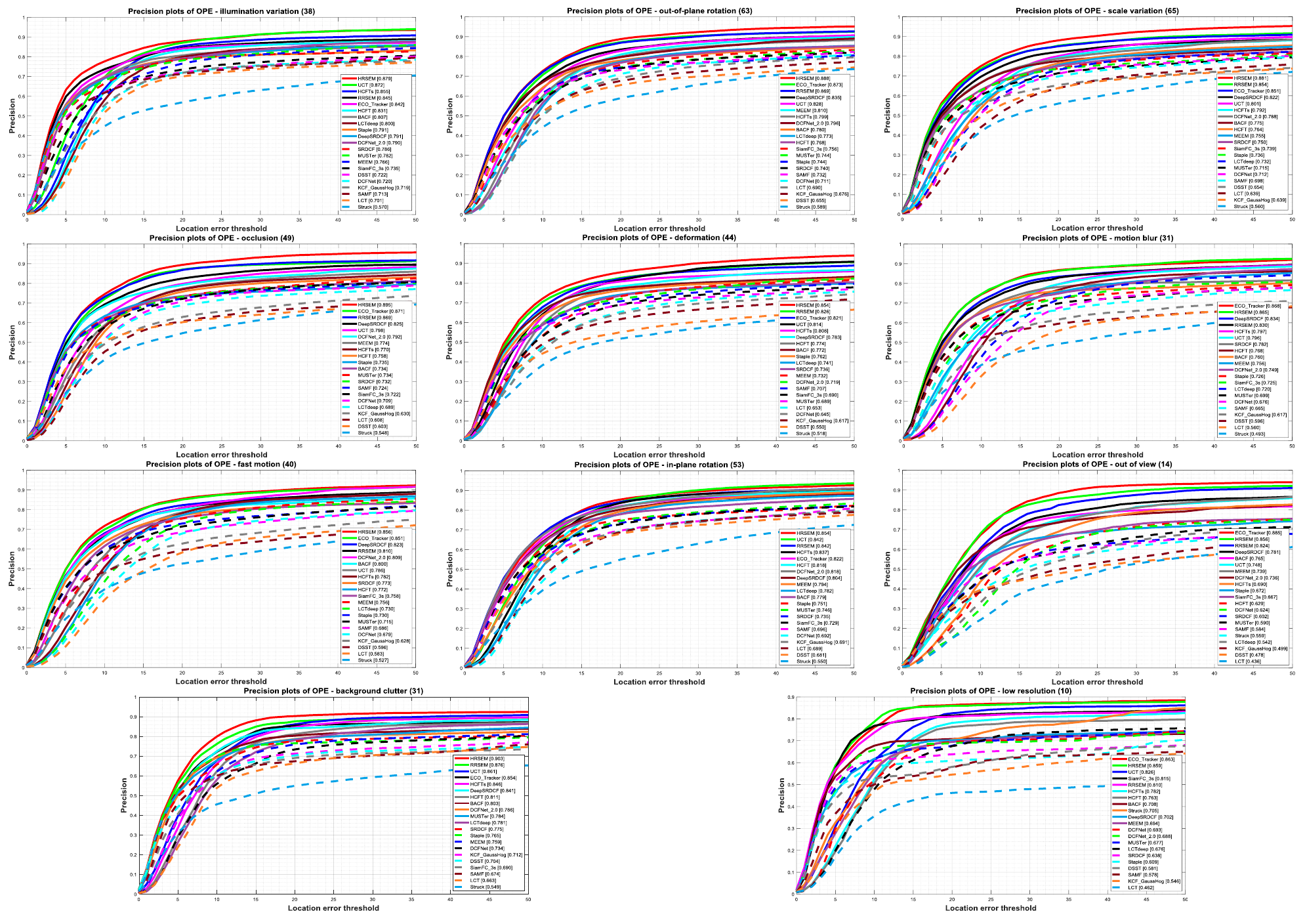}
\caption{Attribute-based evaluations of visual trackers in terms of the average precision rate on OTB-100 dataset.}\label{Fig4_AttComp}
\vspace{-4mm}
\end{figure}

\begin{figure}
\centering
\includegraphics[width=11.9cm, height=6.5cm]{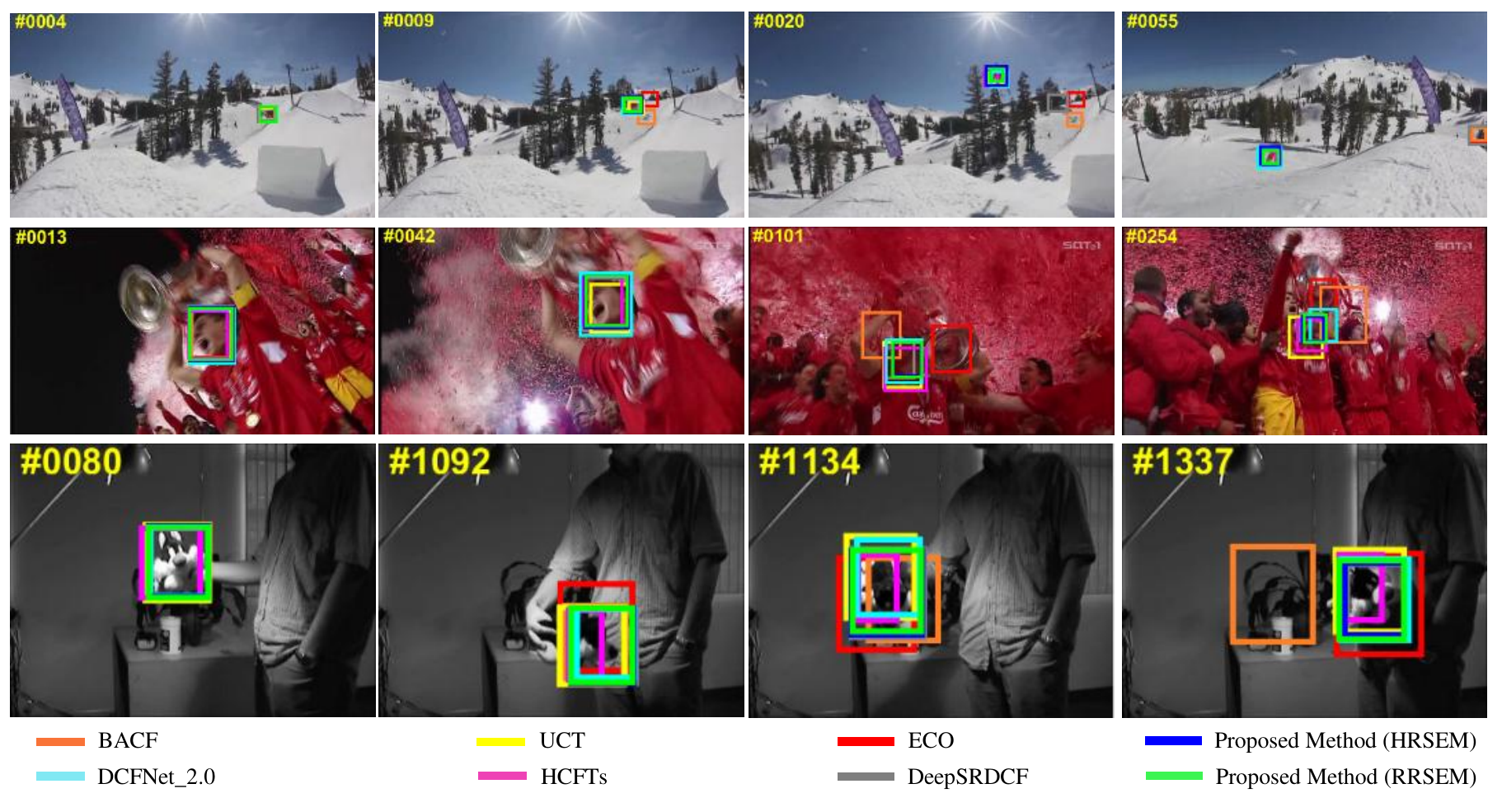}
\caption{Qualitative evaluations of ECO, DeepSRDCF, UCT, HCFTs, BACF, DCFNet-2.0 trackers, and proposed holistic and region representation-based trackers on three video sequences from the OTB-100 dataset (namely: Skiing, Soccer, and Sylvester; from top to bottom row, respectively).}\label{Fig5_QualComp}
\vspace{-4mm}
\end{figure}

The proposed methods lead to competitive results compared to the state-of-the-art methods (e.g., C-RPN \cite{CRPN}, SiamDW \cite{SiamDW}, and LSART \cite{LSART}) that not only have been trained on large-scale datasets for visual tracking purposes but also exploit some strategies such as region proposal networks and detection modules. Moreover, the attribute-based comparisons of the proposed method with state-of-the-art visual trackers on the VOT-2018 and OTB-100 are shown in Fig. \ref{Fig3_VOTComp} and Fig. \ref{Fig4_AttComp}, respectively. According to these results, the proposed methods can achieve more accurate scale estimation for DCF-based visual trackers. Moreover, the qualitative comparisons of the proposed visual tracking methods with top-6 visual trackers including the UCT \cite{UCT}, ECO \cite{ECO}, HCFTs \cite{HCFTs}, DCFNet-2.0 \cite{DCFNet}, MUSTer \cite{MUSTer}, and BACF \cite{BACF} are shown in Fig. \ref{Fig5_QualComp}. 

Based on the obtained results, the proposed HRSEM has achieved the best visual tracking performance at the most challenging attributes compared to the other state-of-the-art visual tracking methods since it can provide a better perception of each visual scene. Also, the proposed formulation effectively integrates into the DCF framework that enjoys the computational efficiency in the Fourier domain.

\section{Conclusion}
\label{sec:5_Conclusion}
Two scale estimation methods based on holistic or region representation of lightweight CNNs were proposed. The formulations of proposed visual trackers exploit the computational efficiency of DCF-based methods in the Fourier domain. The proposed holistic representation-based scale estimation method extracts deep feature maps from full frames and then constructs a scale model according to different scales of the target region. Using a batch of target regions, the proposed region representation-based method extracts 4D deep feature maps to model the target scale, effectively. The comprehensive experimental results on four visual tracking datasets demonstrated that the proposed holistic and region representation-based methods have state-of-the-art visual tracking performance with an acceptable tracking speed. \\

\noindent\textbf{Acknowledgement:} This work was partly supported by a grant (No. $96013046$) from Iran National Science Foundation (INSF).\\

\noindent\textbf{Conflict of interest} All authors declare that they have no conflict of interest. 

\bibliographystyle{plainnat}
\bibliography{ref}

\end{document}